\pdfoutput=1

\documentclass[final,5p,times,twocolumn]{elsarticle}
\usepackage{lineno,hyperref}
\usepackage{algorithmic}
\usepackage{algorithm}
\usepackage{multirow}
\usepackage{subfig}
\usepackage{float}
\usepackage{amsmath}
\usepackage{amssymb}
\usepackage{graphicx}
\modulolinenumbers[5]
\usepackage{booktabs}
\usepackage{color}

\begin{document}

\begin{frontmatter}

\title{Multi-label Learning Based Deep Transfer Neural Network for Facial Attribute Classification}

\author[a1]{Ni Zhuang}
\author[a1]{Yan Yan\corref{corr}}
\ead{yanyan@xmu.edu.cn}
\author[a2]{Si Chen}
\author[a1]{Hanzi Wang}
\author[a3]{Chunhua Shen}

\cortext[corr]{Corresponding author. Tel.:+86-592-2580063}

\address[a1]{Fujian Key Laboratory of Sensing and Computing for Smart City, \\
School of Information Science and Engineering, Xiamen University, Xiamen 361005, China}
\address[a2]{School of Computer and Information Engineering, Xiamen University of Technology, Xiamen 361024, China}
\address[a3]{School of Computer Science, The University of Adelaide, Adelaide, SA 5005, Australia}


\begin{abstract}
Deep Neural Network (DNN) has recently achieved outstanding performance in a variety of computer vision tasks, including facial
attribute classification.
The great success of classifying facial attributes with DNN often relies on a massive amount of labelled data. However, in real-world applications, labelled data are only provided for some commonly used attributes (such as age, gender); whereas, unlabelled data are available for other attributes (such as attraction, hairline). To address the above problem, we propose a novel deep transfer neural network method based on multi-label learning for facial attribute classification, termed FMTNet, which consists of three sub-networks: the Face detection Network (FNet), the Multi-label learning Network (MNet) and the Transfer learning Network (TNet). Firstly, based on the Faster Region-based Convolutional Neural Network (Faster R-CNN), FNet is fine-tuned for face detection. Then, MNet is fine-tuned by FNet to predict multiple attributes with labelled data, where an effective loss weight scheme is developed to explicitly exploit the correlation between facial attributes based on attribute grouping.
Finally, based on MNet, TNet is trained by taking advantage of unsupervised domain adaptation for unlabelled facial attribute classification. The three sub-networks are tightly coupled to perform effective facial attribute classification. A distinguishing characteristic of the proposed FMTNet method is that the three sub-networks (FNet, MNet and TNet) are constructed in a similar network structure. Extensive experimental results on challenging face datasets demonstrate the effectiveness of our proposed method compared with several state-of-the-art methods.
\end{abstract}
\end{frontmatter}
\section{Introduction}
Facial attribute classification is an important and fundamental research area in computer vision and pattern recognition. The task of facial attribute classification is to predict the attributes of a facial image, including gender, attraction, race, etc. Recently, facial attribute classification has received increasing attention with a wide range of applications, such as face verification \cite{sun2015deeply3,kumar2009attribute4,kumar2011describable8}, face recognition \cite{chen2014cross-age1,yu2017discriminative,li2016robust}, face retrieval \cite{chen2015face2}. However, it remains a challenging problem, because of the large facial appearance variations caused by pose, illumination and occlusion, etc.\\
\indent Early works on facial attribute classification usually characterize the facial attributes based on the histogram representation \cite{kumar2009attribute4,kumar2011describable8,cherniavsky2010semi-supervised9}. For example, Kumar \textit{et al.}~\cite{kumar2009attribute4} propose to firstly extract the low-level features from different regions of a face, and then predict facial attributes with the Support Vector Machine (SVM) for face verification. Cherniavsky \textit{et al.}~\cite{cherniavsky2010semi-supervised9} develop a generative facial feature representation method based on the Haar-like features and investigate a semi-supervised method to predict facial attributes with SVM.\\
\indent Recent research mainly focuses on using the Deep Neural Network (DNN) to predict facial attributes. Luo \textit{et al.}~\cite{luo2013a10} combine discriminative decision trees with the deep Sum-Product Network (SPN) for facial attribute classification. In \cite{zhang2014panda11,liu2015deep12,kang2015face13}, the authors firstly extract facial features using DNN and then classify  facial attributes with SVM. Ehrlich \textit{et al.}~\cite{ehrlich2016facial} learn the shared feature representation for facial attributes by directly operating on faces and facial landmark points. Rudd \textit{et al.}~\cite{rudd2016moon} address the problem of imbalanced data to predict multiple facial attributes.\\
\indent Generally speaking, methods for facial attribute classification can be divided into two categories: single-label learning based methods \cite{kumar2009attribute4,cherniavsky2010semi-supervised9,zhang2014panda11,liu2015deep12,kang2015face13} and multi-label learning based methods \cite{ehrlich2016facial,rudd2016moon}. The single-label learning based methods predict facial attributes separately and thus do not consider the correlation between facial attributes. In contrast, the multi-label learning based methods, which attempt to predict facial attributes simultaneously by using labelled data, have drawn increasing attention. However, in real-world applications, only some commonly used attributes are provided with labelled information, while the other attributes have unlabelled data. Therefore, these methods \cite{ehrlich2016facial,rudd2016moon} fail to deal with the facial attribute classification problem when unlabelled information is available (recall that these methods are based on supervised learning).\\
\indent Motivated by the above observations, we propose a novel facial
\begin{figure*}[!t]
\centering
\centerline{\includegraphics[width=1\textwidth,height=0.7\textwidth]{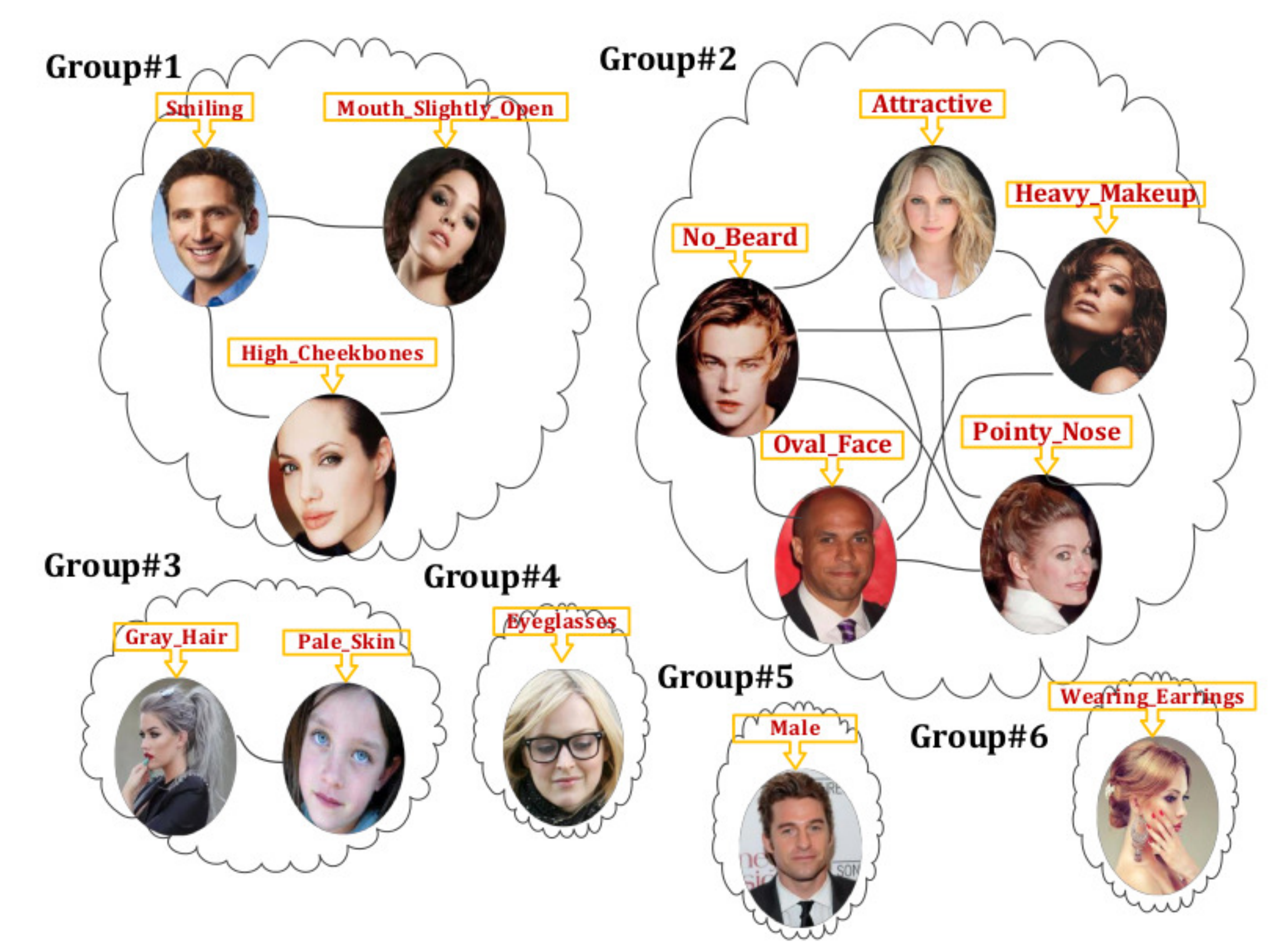}}
\caption{The correlation between facial attributes. Based on the work of Liu \textit{et al.}~\cite{liu2015deep12}, we obtain the correlation between different attributes. The facial  attributes with high correlation can be clustered into the same group.}
\label{fig:1}
\end{figure*}
attribute classification method, which performs transfer learning based on multi-label learning. More specifically, we take advantage of the transfer DNN technique to predict facial attributes that do not have labelled information in the target domain. To effectively exploit the labelled data in the source domain, we use the multi-label learning technique to predict multiple facial attributes simultaneously, considering the correlation between facial attributes. Fig.~1 shows an illustration of the correlation between different facial attributes. For different learning problems, some carefully designed networks are used, where these networks share the same structure at the former layers of the networks and they only differ at the latter layers. Therefore, the networks can be effectively trained via fine-tuning.\\
\indent In this paper, we propose an effective deep transfer neural network method, termed FMTNet, which consists of three sub-networks for facial attribute classification. The first sub-network is the Face detection Network (FNet) for face detection. FNet is initialized by using the model learned from a large scale ImageNet dataset \cite{deng2009imagenet31}, and then is fine-tuned by using the facial images. The second sub-network is the Multi-label learning Network (MNet) for facial attribute classification with supervised learning, where multiple attributes are predicted simultaneously. Based on FNet, MNet is fine-tuned by using labelled attributes in the source domain. The network structures at the former layers of both MNet and FNet are the same, whereas the main difference is that multiple fully-connected layers are independently constructed in MNet. The third sub-network is the Transfer learning Network (TNet) for facial attribute classification, when labelled information is not available in the target domain. Based on MNet, TNet makes use of unsupervised domain adaptation to improve the performance of facial attribute classification.\\
\indent The main contributions of this paper are summarized as follows:\\
\indent (1) Instead of using single-label learning for each attribute \cite{chen2014cross-age1,chen2015face2}, the proposed method effectively performs facial attribute classification based on multi-label learning for the labelled attributes in the source domain. Especially, we propose an effective loss weight scheme to explicitly exploit the correlation between facial attributes based on attribute grouping, which can significantly improve the generalization performance of the proposed method.\\
\indent (2) Based on multi-label learning, the proposed method leverages transfer learning to predict facial attributes for the unlabelled attributes in the target domain. The transfer neural network successfully transfers the  features from the source domain (with labelled information) to the target domain (without labelled information), even when the probability distributions between the two domains are significantly different. Therefore, the proposed method alleviates the dependency on fully labelled training data, especially in the absence of labelled information for some attributes. \\
\indent The remainder of the paper is organized as follows: In Section 2, some related work is discussed. In Section 3, the details of the proposed FMTNet method for facial attribute classification are described.  In Section 4, the experimental results are reported. In Section 5, the conclusions are presented.
\section{Related Work}
The proposed method is closely related to deep learning, multi-label learning and transfer learning. In this section, we briefly discuss the related work.

\subsection{Deep Learning}
The recent great success of deep learning based facial attribute classification is triggered by a growing number of works \cite{zhang2014panda11,liu2015deep12,kang2015face13,zhong2016leveraging14,Luo201615} on learning compact and discriminative features. Compared with the traditional feature extraction methods (such as Scale-Invariant Features Transform (SIFT) \cite{lowe2004distinctive16} and Histogram of Oriented Gradients (HOG) \cite{dalal2005histograms17}), the deep learning based methods \cite{zhang2014panda11,liu2015deep12,zhong2016leveraging14} have shown astounding performance improvement. For example, Zhang \textit{et al.}~\cite{zhang2014panda11} combine the deep features obtained from each poselet of the face region with the deep features of the whole facial image as the final features, and then use SVM for facial attribute classification. Liu \textit{et al.}~\cite{liu2015deep12} employ DNN for face localization and apply another neural network to extract features in the face region for attribute classification. Zhong \textit{et al.} \cite{zhong2016leveraging14} extract the features of each attribute based on the hierarchical DNN, and use the linear SVM to classify facial attributes. All the above methods rely on DNN to extract features and require an additional classifier to classify facial attributes. Furthermore, they usually do not consider the correlation between facial attributes. Different from these methods, our proposed method presents an end-to-end network, which effectively exploits the correlation between facial attributes, thus considerably improving the performance of facial attribute classification.
\subsection{Multi-label Learning}
Multi-label learning \cite{zhang2014review,huang2013multi27,pillai2017designing,wang2016dynamic,li2016facial}, which aims to learn multiple different but related labels simultaneously, has received much attention so far. For example, Xu \textit{et al.} \cite{Xu2016Local} analyze the local Rademacher complexity of empirical risk minimization (ERM)-based multi-label learning algorithms and then propose a new method that not only results in a sharp generalization error bound, but also provides a tight approximation of the low-rank structure. In \cite{Xu2016Robust}, the authors present an additional sparse component to deal with the tail labels for the multi-label learning task. In \cite{You2017Privileged}, You \textit{et al.} present the privileged multi-label learning (PrML) method to exploit the correlation between labels.
For the problem of multi-label learning with missing labels, Jain \textit{et al.} \cite{JainScalable} develop a scalable and generative framework, which is based on a latent factor model for the label matrix and an exposure model for missing labels. \\
\indent Recently, the DNN based multi-label learning method has been proposed for facial attribute classification in \cite{rudd2016moon}.
The features learned by DNN exhibit the hierarchical structure (such as pixels, edges, object parts and objects), where the low-level features usually share the similar representation. Therefore, the multi-label learning based on DNN uses the shared features at the low-level layers and separately learns the features of multiple facial attributes at the high-level layers. However, such a method focuses on the imbalance problem of the data and does not explicitly consider the correlation between facial attributes.
In this paper, we decompose the multi-label learning problem into a number of binary classification problems, where we solve all these problems using a single neural network at once.
Furthermore, we not only consider multiple facial attributes learning, but also investigate the relationship between the different attributes based on attribute grouping.
\subsection{Transfer Learning}
The general performance of the traditional classifier trained with a limited number of labelled data may not be satisfactory, while the manual annotation of abundant training data for various tasks costs too much manpower. Fortunately, transfer learning \cite{pan2010survey37,wei2016facial,hu2017sparse} is an effective technique to improve the performance of the classifier in the target domain given only the annotated data in the source domain, which greatly reduces the labelling cost. Transfer learning also refers to unsupervised domain adaptation \cite{long2013transfer24,long2014transfer25,yosinski2014transferable26}, and it can adapt the features from the labelled source domain to the unlabelled target domain. Long \textit{et al.} \cite{long2013transfer24} consider both the marginal and conditional distributions between the source and target domains via a Joint Distribution Adaptation (JDA) method. In \cite{long2014transfer25}, the authors propose to reduce the domain difference by jointly matching the features and reweighing  the instances via a Transfer Joint Matching (TJM) method. Han \textit{et al.}~\cite{Han2010Multi} propose a sparse multi-label transfer learning framework, which first learns a multi-label encoded sparse linear embedding space, and then maps the target data onto the learned space for circumventing the problem of multiple tags in practical applications. \\
\indent With the development of deep learning, the DNN based transfer learning methods \cite{long2016unsupervised22,long2015learning23} usually add the adaptation layers to effectively reduce the discrepancy between the source domain and the target domain. For example, Long \textit{et al.}~\cite{long2015learning23} generalize DNN to domain adaptation, which is formulated as the Deep Adaptation Network (DAN). To the best of our knowledge, our proposed method is the first attempt for the domain adaptation of facial attribute classification. More importantly, we propose to use the deep transfer neural network based on multi-label learning for facial attribute classification.

\section{The Proposed Method}
The proposed facial attribute classification method (i.e., FMTNet) mainly consists of three sub-networks: 1) the Face detection Network (FNet) for face detection; 2) the Multi-label learning Network (MNet) for predicting multiple facial attributes simultaneously; and 3) the Transfer learning Network (TNet) for unlabelled facial attribute classification. These three sub-networks are designed to share the same convolutional layers (i.e., the first 13 convolutional layers in VGG-16 \cite{simonyan2014very29}), so that they can be easily fine-tuned. Fig.~2 shows the overall framework of the proposed FMTNet method for facial attribute classification. A detailed description of the proposed method is given in the following subsections.
\begin{figure*}[!t]
\centering
\centerline{\includegraphics[width=7in,height=3in]{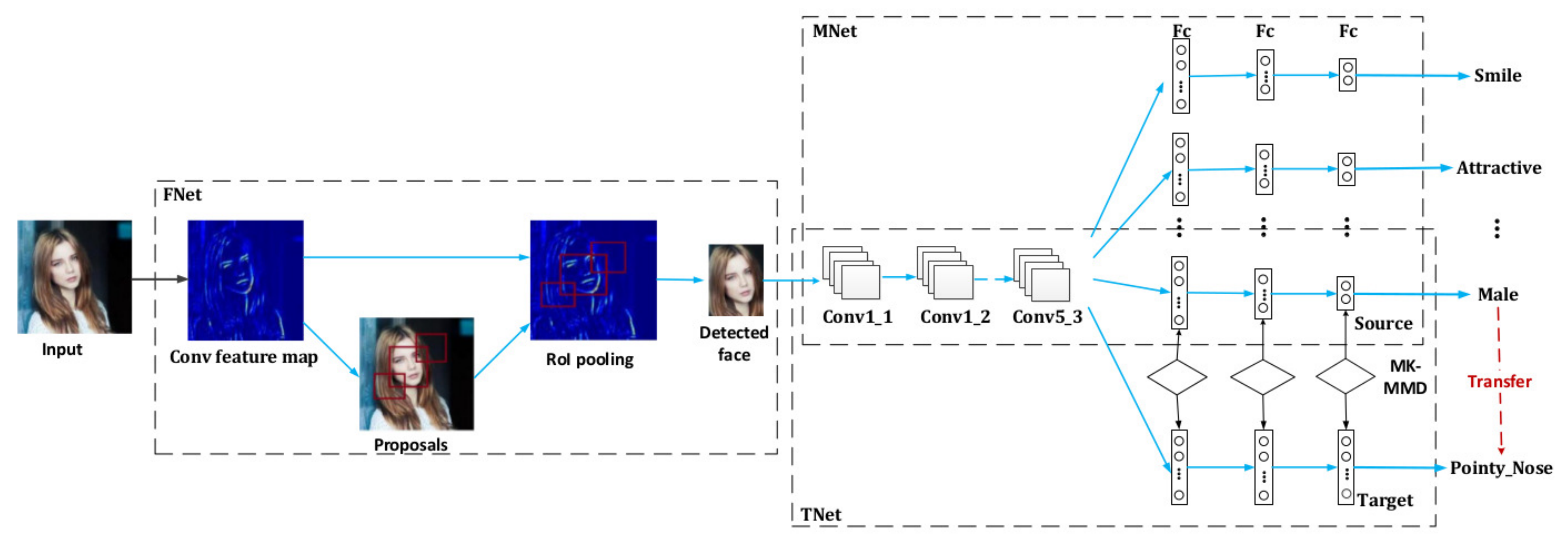}}
\caption{The overall framework of the proposed FMTNet method for facial attribute classification. The first sub-network is FNet for face detection. The second sub-network is MNet for facial attribute classification with supervised learning, where multiple facial attributes are predicted simultaneously. The third sub-network is TNet for facial attribute classification when labelled data are not available. The three sub-networks share the same structure at the former layers but they  differ at the latter layers.}
\label{fig:2}
\end{figure*}

\subsection{FNet (Face detection Network)}
FNet is designed to perform face detection, which outputs the positions of the faces. In this paper, we mainly follow the work of Faster R-CNN \cite{ren2015faster} to train FNet. Specifically, as shown in Fig.~2, we first take an image (of any size) as the input and use VGG-16 \cite{simonyan2014very29}, which has 13 shareable convolutional layers, to obtain the feature map of the whole image. Then, we use the Region Proposal Network (RPN) to obtain a set of rectangular face proposals, each of which has a face score. The RoI pooling layer combines the convolutional features with the predicted bounding boxes, so that FNet can process a theoretical valid back-propagation. Finally, we get the face regions with reference to a series of pre-defined boxes by using the box regression.\\
\indent The above training scheme shows the discriminative capability of FNet for face detection. Most importantly, FNet is used as the basis of MNet and TNet (i.e., these three sub-networks share the same network structures for their former 13 convolutional layers).
\subsection{MNet (Multi-label learning Network)}
The second sub-network is the Multi-label learning Network (MNet), which simultaneously predicts multiple facial attributes with labelled data. Specifically, we fine-tune the former 13 convolutional layers (which are the same as those in FNet) and train the latter three fully-connected layers for each attribute, as shown in Fig.~2. To reduce the model complexity of MNet, we decrease the dimensions of two fully-connected layers used in VGG-16 \cite{simonyan2014very29}. Specifically, we reduce the dimensions of the `fc14' fully-connected layer from 4,096 to 512, and the dimensions of the `fc15' fully-connected layer from 4,096 to 256. As shown in the experiments, we still obtain on par results with the lower dimensions of the fully-connected layers, thus reducing computational complexity. \\
\indent MNet predicts one attribute in conjunction with the other attributes. The structure of MNet (see Fig.~2) contains two parts: (1) the shared layers which are collaboratively trained for all the attributes, and (2) the independent layers which are separately trained for each attribute. In this paper, we tackle the multi-label learning in an attribute-by-attribute style and decompose the multi-label learning problem into a number of binary classification problems. Moreover, in order to improve the generalization performance of MNet, we exploit the correlation between facial attributes based on attribute grouping.\\
\indent Assume that there are $\textit{I}$ facial attributes to be predicted. Let us define the softmax loss term, $\textit{$L_{i}$}$, associated with the $\textit{i}$-th attribute, and the corresponding class label (denoted as $\textit{$y_{i}$}$) is written as follows:
\begin{equation}
L_{i} = -\frac{1}{N}\sum\limits_{n=1}^{N}log(p_{n,y_{i}}),
\end{equation}
where
\begin{equation}
p_{n,y_{i}}=\frac{e^{x_{n,y_{i}}}}{\sum\limits_{y_{i}=1}^{C_{i}}e^{x_{n,y_{i}}}},
\end{equation}and $\textit{N}$ is the number of training examples; $\textit{$p_{n,y_{i}}$}$ represents the probability of the $\textit{i}$-th attribute class calculated by the softmax function; $\textit{$x_{n,y_{i}}$}$ represents the value of the last fully-connected layer predicted in the $\textit{i}$-th attribute class; and $\textit{$y_{i}$}$ can take on $\textit{$C_{i}$}$ values, ranging from 1 to $\textit{$C_{i}$}$.\\
\indent In this paper, we consider the facial attribute classification problem as a binary classification problem (i.e., $\textit{$C_{i}$}$ = 2). In other words, an example, which owns the corresponding label, is treated as a positive sample (i.e., $\textit{$y_{i}$}$ = 1); otherwise, it is considered a negative sample (i.e., $\textit{$y_{i}$}$ = 2). Of course, each facial attribute classification on MNet is not limited to the binary classification, and it can be easily extended to multi-class classification.\\
\indent Thus, we derive the loss function of MNet as follow:
\begin{equation}
L = \sum\limits_{i=1}^{I} \lambda_{i}L_{i},
\end{equation}
where $\textit{$\lambda_{i}$}$ denotes the loss weight for the $\textit{i}$-th attribute. \\
\indent In MNet, multiple attributes are trained together, where each attribute is associated with a loss weight (see Eq.~(3)). The loss weight has significant influence on the performance of facial attribute classification (see Section 4.2.1 for the experimental results). Moreover, Liu $\textit{et al.}$ \cite{liu2015deep12} show that the facial attributes have clear grouping patterns. In other words, the facial attributes can be clustered into several groups, where the attributes in one group show high correlation and those between groups have low correlation (see Fig.~1). Therefore, based on this observation, we propose to define the loss weight for each attribute by taking advantage of attribute grouping, which effectively considers the relationship between different attributes. The details of the proposed loss weight scheme are given as follows.

\indent Firstly, we cluster all the \textit{I} facial attributes into \textit{G} groups (we use the clustering method in \cite{liu2015deep12}). Suppose that there are $\textit{$g_{m}$}$ attributes (and their corresponding loss weights are defined as $\textit{$\lambda_{g_1}$}$, $\textit{$\lambda_{g_2}$}$, ..., $\textit{$\lambda_{g_m}$}$) for the $\textit{g}$-th group. The  loss weights for all the $\textit{$g_{m}$}$ attributes are set to be the same, and their sum is equal to \textit{1/G}. Therefore,
\begin{equation}
\lambda_{g_1} = \lambda_{g_2} = \cdots = \lambda_{g_m}=\frac{1}{G}\cdot\frac{1}{g_m}.
\end{equation}
If there is only one attribute in a group, we directly set its corresponding loss weight to be \textit{1/G}. For example, we can assume that 6 facial attributes are clustered into 3 groups, where one group (Group 1) has 2 attributes and the other two groups (Group 2 and Group 3) have 3 attributes and 1 attribute, respectively. In this case, the loss weight for each attribute in Group 1 is $1/3\times 1/2=1/6$, while that for each attribute in Group 2 is $ 1/3\times 1/3=1/9$. The loss weight for the attribute in Group 3 is $1/3$. Thus, the sum of the loss weights for the attributes in each group is the same (i.e., $1/3$). In other words, the proposed loss weight scheme can effectively balance the loss weight of each group (no matter how many attributes are in this group) and prevents TNet from overfitting to one group (especially when many attributes exist in that group). As a result, the performance of MNet can be improved.\\
\indent Subsequently, we train MNet by using the stochastic gradient descent (SGD) algorithm to minimize the loss function on the training data.\\
\indent Compared with traditional single-label learning methods \cite{kumar2009attribute4,cherniavsky2010semi-supervised9,zhang2014panda11,liu2015deep12,kang2015face13}, which independently train different attributes, MNet effectively overcomes this problem by training multiple attributes simultaneously. Accordingly, the shared layers in MNet significantly reduce the number of network parameters (recall that the DNN based single-label learning methods need to estimate the network parameters for each attribute).\\
\indent Different from the DNN based multi-label learning method \cite{rudd2016moon}, the proposed MNet successfully exploits the correlation between these attributes based on attribute grouping. Recall that we cluster the attributes, which have high correlation, into one group, and then assign the loss weights according to different attribute groups. Therefore, the proposed loss weight scheme takes advantage of attribute grouping to assign the loss weights. It is worth pointing out that attribute grouping is a key step in the proposed method. This is because that not only the proposed loss weight scheme in MNet is based on attribute grouping, but also the performance of TNet is greatly affected by the results of attribute grouping (see Section 4.3.3).
\subsection{TNet (Transfer learning Network)}
The third sub-network is the Transfer learning Network (TNet), which explores the transferability of one attribute with labelled information to another attribute without labelled information. To optimize the training process, TNet shares the similar network structure as MNet and it is fine-tuned based on MNet. In transfer learning, the domain with labelled information is treated as the source domain, and the domain without labelled information is considered as the target domain. Moreover, data in these two domains are usually under different probability distributions. As discussed previously, the attributes in the source domain are trained using the multi-label learning method. In this paper, to deal with different data distributions in the two domains, we introduce a Reproducing Kernel Hilbert Space (RKHS) that is a high-dimensional space, where the domain discrepancy is measured by using the Multi-Kernels Maximum Mean Discrepancies (MK-MMD) criterion proposed by Gretton \textit{et al.}~\cite{gretton2012optimal5}. Fig.~3 gives an intuitive example of transfer learning.\\
\begin{figure*}[!t]
\centering
\centerline{\includegraphics[width=0.95\textwidth,height=0.3\textwidth]{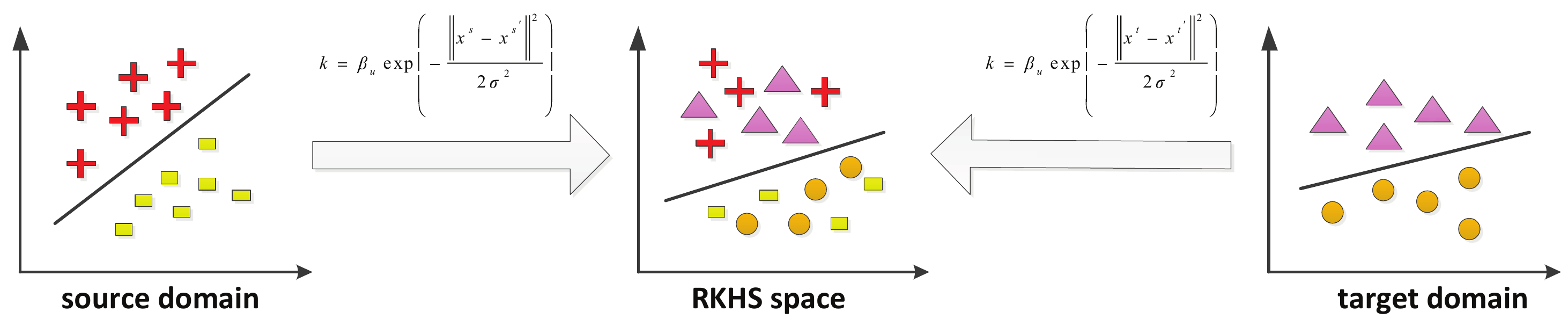}}
\caption{An intuitive example about transfer learning. We use the kernel function to map the samples from the source domain and the target domain into the RKHS space, which can make the two domains closer. In the source domain, the red crosses and the yellow strips denote the positive samples and the negative samples, respectively. In the target domain, the pink triangles and the wheat circles denote the positive samples and the negative samples, respectively. }
\label{fig:3}
\end{figure*}
\indent Assume that $X^{s}$ = $\lbrace x^{s}_{1}, x^{s}_{2}, ..., x^{s}_{m} \rbrace$ consists of $\textit{m}$ data with the labelled information $Y^{s}$ in the source domain, and $X^{t}$= $\lbrace x^{t}_{1}, x^{t}_{2}, ..., x^{t}_{n} \rbrace$ consists of $\textit{n}$ data in the target domain. Thus, $\textit{$D_{s}=\lbrace(X^{s},Y^{s})\rbrace$}$ represents the source domain, and $\textit{$D_{t}=\lbrace(X^{t})\rbrace$}$ represents the target domain. Furthermore, let $\textit{p}$ denote the probability distribution of the source domain and $\textit{q}$ denote the probability distribution of the target domain. $\textit{$H_{k}$}$ is assumed to be a RKHS defined on a topological space $\textit{$\chi $}$ with the characteristic kernel $\textit{k}$. The MK-MMD $\textit{$d_{k}(p,q)$}$ is defined as the distance between the probability distributions $\textit{p}$ and $\textit{q}$ in the RKHS. Therefore, we formulate the squared form of MK-MMD as follows:
\begin{equation}
d_{k}^{2}(p,q)= \|E_{p}[\phi(X^{s})]-E_{q}[\phi(X^{t})]\|_{H_{k}}^{2},
\end{equation}
where $\textit{$E_{p}[\cdot]$}$ is the mean of $\textit{p}$ and $\textit{$E_{q}[\cdot]$}$ is the mean of \textit{q}. \textit{$\phi(\cdot)$} is a feature mapping function, which maps the features from the original feature space to the RKHS space \cite{gretton2012optimal5}.\\
\indent In MK-MMD, we select a kernel from a particular family of kernels $\textit{K}$ (here, the multiple Gaussian kernels are used for optimal kernel selection). Let us denote $\textit{$\lbrace k_{u}\rbrace _{u=1}^{d}$}$ as a set of positive definite functions, where $\textit{$k_{u}$}$: $\textit{$\chi \times \chi \rightarrow R$}$. Hence,
\begin{equation}
K=\lbrace k : k=\sum\limits_{u=1}^{d}\beta_{u}k_{u},\sum\limits_{u=1}^{d}\beta_{u}=1, \beta_{u}\geqslant0, \forall u \in \lbrace1, ..., d\rbrace \rbrace,
\end{equation}
where $\textit{$\beta_{u}$}$ is the coefficient of the \textit{u}-th kernel. We constrain $\textit{$\beta_{u}$}$ to guarantee that the derived kernel is characteristic.\\
\indent The characteristic kernel, \textit{$k(\cdot)$}, is defined as $\textit{$k(x^{s},x^{t}) = <\phi(x^{s}),\phi(x^{t})>$}$.~Accordingly, the distance between two probability distributions is computed as follows:
\begin{align}
\nonumber
d_{k}^{2}(p,q)&=E_{x^{s},x^{s{'}}}[k(x^{s},x^{s{'}})]-2E_{x^{s},x^{t}}[k(x^{s},x^{t})]\\
  &+E_{x^{t},x^{t{'}}}[k(x^{t},x^{t{'}})],
\end{align}
where $x^{s}$, $x^{s{'}}$ are two features in the source domain and $x^{t}$, $x^{t'}$ are two features in the target domain. $\textit{$E_{x^{s},x^{s{'}}}[\cdot]$}$ is the mean of $\textit{$x^{s}$}$ and $\textit{$x^{s{'}}$}$, and $\textit{$E_{x^{s},x^{t}}[\cdot]$}$ and $\textit{$E_{x^{t},x^{t{'}}}[\cdot]$}$ are similarly defined.\\
\indent The objective of transfer learning is to minimize the domain discrepancy between the source domain and the target domain, which is effectively measured by the distance between the probability distributions from the source domain and the target domain (i.e., MK-MMD). Therefore, we have:
\begin{equation}
\min_{D_{s},D_{t}}D_{F}(D_{s},D_{t})= \min_{p,q}d_{k}^{2}(p,q),
\end{equation}
where \textit{$D_{F}$} ($=d_{k}^{2}(p,q)$) denotes the domain discrepancy between two domains in the fully-connected layer.\\
\indent We respectively embed three MK-MMD loss layers in the last three layers of TNet, and thus the total loss of the whole TNet is the sum of the softmax loss and three MK-MMD losses, which can be written as:
\begin{equation}
L=\sum\limits_{i=1}^{3} {D_{F_{i}}}+ \alpha L_{s},
\end{equation}
where \textit{L} is the total loss of the whole TNet, which is trained by the SGD method. The ${D_{F_{i}}}$ denotes the MK-MMD loss in the last \textit{i}-th layer, and $L_{s}$ represents the softmax loss of the source domain ($\alpha$ is the loss weight).\\
\indent The loss of TNet (see Eq.~(9)) consists of two terms (i.e., the MK-MMD loss and the softmax loss). The objective of minimizing the MK-MMD loss is to reduce the discrepancy between the distributions of the source and target domains, while the objective of minimizing the softmax loss is to enhance the discriminability of the features in the source domain. In this paper, the MK-MMD loss is based on the kernel mean matching, where only features of data are used (while the labeled information is not used). Meanwhile, the softmax loss is defined in the source domain, where the labelled information is considered. By combining these two terms (i.e., the MK-MMD loss and the softmax loss), the discriminability of the features in the target domain is enhanced.\\
\indent An example is given in Fig.~2, where the features learned from the `Male' attribute are transferred to the `Pointy\_Nose' attribute. The `Male' attribute has a large amount of labelled training data in the source domain; whereas, the `Pointy\_Nose' attribute has unlabelled training data in the target domain. Although the network structures of these two attributes are the same, the discrepancy of the features obtained in the latter layers is large. As shown in the experiments, the performance of feature transferability significantly drops if we perform direct transfer (i.e., train the network with the source data having labelled information, and then directly extract the features of the target data for classification). However, based on the MK-MMD loss function, the domain discrepancy between the `Male' attribute and the `Pointy\_Nose' attribute is effectively measured. As a result, by minimizing the MK-MMD loss, the difference between the two attributes is greatly reduced so that the feature transferability is significantly improved.\\
\indent It is worth pointing out that TNet and MNet share the same feature extraction layers (i.e., the former 13 convolutional layers), while the latter fully-connected layers are different (see Fig.~2 for an illustration). Considering the limited number of training data, TNet directly uses the parameters of the first 6 convolutional layers, while fine-tuning the parameters of the subsequent 7 convolutional layers of MNet (i.e., we freeze `conv1\_1' - `conv3\_2' to learn generic features and fine-tune `conv3\_3' - `conv5\_3' to correct the slight domain biases on the convolutional layers of VGG-16 \cite{simonyan2014very29}). This is because that the fine-tuning strategy is usually much easier to obtain the network parameters which are learned by taking advantage of multi-label learning based on the labelled facial attributes data.\\
\indent Although recent works \cite{long2016unsupervised22,long2015learning23}
learn the transferable features with DNN, which also uses the MK-MMD loss function, the proposed TNet has some differences: 1) The proposed TNet focuses on facial attribute classification, while the methods in \cite{long2016unsupervised22,long2015learning23} are proposed for office image classification. 2) The proposed TNet is fine-tuned based on MNet and thus the network structures of TNet and MNet are similar. As a result, TNet takes both advantages of transfer learning and multi-label learning. In contrast, the methods in \cite{long2016unsupervised22,long2015learning23} mainly perform transfer learning via the adaptation network.
\subsection{The Overall FMTNet}
In the previous subsections, we have developed all the ingredients for the proposed FMTNet method. Specifically, we use FNet for face detection, use MNet for predicting multiple facial attributes simultaneously with labelled data, and use TNet for predicting facial attributes without labelled information. The training process of the overall FMTNet method is described in algorithm 1.
\begin{algorithm}[!h]
\caption{The training process of the proposed FMTNet method}
\label{alg:1}
\begin{algorithmic}[1]
\renewcommand{\algorithmicrequire}{ \textbf{Input:}}
\renewcommand{\algorithmicensure}{ \textbf{Output:}}
\REQUIRE
The training datasets $\mathcal{D}_{f}$, $\mathcal{D}_{m}$, and $\mathcal{D}_{t}$ for FNet, MNet and TNet, $~~$~~~~~  respectively.
\ENSURE  The network parameters $\mathcal{W}_{f}$, $\mathcal{W}_{m}$, and $\mathcal{W}_{t}$ for FNet, MNet and $~~$~~~~~  TNet,  respectively.
\STATE \textbf{Initialization:} Initialize $\mathcal{W}_{f}$ with VGG-16;
\WHILE{not converge}
  \FOR{each training sample $\textbf{x}_{i}  \in \mathcal{D}_{f}$ }
     \STATE Forward pass to obtain the feature representation of $\textbf{x}_{i}$;
     \STATE Back propagate to update the network parameters $\mathcal{W}_{f}$;
  \ENDFOR
\ENDWHILE
\STATE Cluster all the attributes into $G$ groups;
\STATE Calculate the loss weight for each attribute via Eq.~(4);
\STATE \textbf{Initialization:} Initialize $\mathcal{W}_{m}$ with $\mathcal{W}_{f}$;
\WHILE{not converge}
  \FOR{each training sample $\textbf{x}_{i}  \in \mathcal{D}_{m} $ }
     \STATE Forward pass to obtain the feature representation of $\textbf{x}_{i}$;
     \STATE Back propagate to update the network parameters $\mathcal{W}_{m}$  via Eq.~(3);
  \ENDFOR
\ENDWHILE
\STATE \textbf{Initialization:} Initialize $\mathcal{W}_{t}$ with $\mathcal{W}_{m}$;
\WHILE{not converge}
  \FOR{each training sample $\textbf{x}_{i}  \in \mathcal{D}_{t} $ }
     \STATE Forward pass to obtain the feature representation of $\textbf{x}_{i}$;
     \STATE Back propagate to update the network parameters $\mathcal{W}_{t}$ via Eq.~(9);
  \ENDFOR
\ENDWHILE
\end{algorithmic}
\end{algorithm}
\section{Experiments}
In this section, we show the effectiveness of the proposed method for facial attribute classification. In Section 4.1, the datasets and parameter settings are described. In Section 4.2, the performance of MNet is evaluated.
The influence of loss weight is given in Section 4.2.1. The comparison between single-label learning and multi-label learning is shown in Section 4.2.2. Performance comparison with several state-of-the-art multi-label learning methods is described in Section 4.2.3.
Finally, in Section 4.3, the performance of the proposed FMTNet is evaluated. The performances obtained by direct transfer and transferring under different correlations are given in Sections 4.3.1 and 4.3.2, respectively. Performance comparison with several state-of-the-art transfer learning methods is described in Section 4.3.3.

\subsection{Datasets and Parameter Settings}
We train and evaluate different facial attribute classification methods on the CelebA \cite{sun2014deep7} and LFWA \cite{huang2007labeled32} datasets.
The CelebA dataset has 202,599 facial images of 10,177 identities, and 40 binary attribute annotations are provided for each facial image. The CelebA dataset is divided into three parts:  training, validation and test. More specifically, the training set, the validation set and the test set respectively contain 162,770 images, 19,867 images and 19,962 images.
The LFWA dataset, with more than 1,680 identities, contains more than 13,000 facial images collected from the web. Each image in this dataset has 73 binary attribute annotations, where the positive or negative values indicate the presence or absence of the corresponding attributes, respectively. We use half of the LFWA dataset for training and half of the LFWA dataset for test. \\
\indent Recall that the proposed method focuses on facial attribute classification,  where the detected faces are used to perform subsequent classification. Therefore, we do not compare FNet with other face detection methods in this paper. In fact, the performance of the Faster R-CNN for face detection is reported in \cite{jiang2016face}. In this experiment, we use the training data from the CelebA dataset to train FNet and then use FNet as the basis of both MNet and TNet.

\begin{table*}[!t]
\caption{The classification accuracy (\%) obtained by MNet with different loss weight schemes. `MNet\_Eq', `MNet\_Emp' and `MNet\_Prop' represent MNet with the same loss weight scheme, MNet with the emphasized loss weight scheme and MNet with the proposed loss weight scheme, respectively. The best results are in boldface.}
\center
\scalebox{0.8}{
\begin{tabular}{c|c|c|c|c|c|c}
\hline
\multirow{3}{*}{Attributes} &
\multicolumn{3}{c|}{CelebA} &
\multicolumn{3}{|c}{LFWA}\\
\cline{2-7}
  & MNet\_Eq & MNet\_Emp & MNet\_Prop & MNet\_Eq & MNet\_Emp & MNet\_Prop\\
\hline
High\_Cheekbone & 87.03 & 87.04 &\textbf{88.19} & 84.29&82.34 &\textbf{85.79}\\
\hline
Mouth\_Open & 93.72 & 93.59 &\textbf{94.16} & 77.92&75.27 &\textbf{81.59}\\
\hline
Smiling & 92.39 & 92.51 &\textbf{93.21} & 88.57&83.68 &\textbf{89.49}\\
\hline
Attractive & 82.46 & 83.09 & \textbf{83.29} & 73.93& 77.59&\textbf{77.78}\\
\hline
Bangs & 95.94 & \textbf{96.08} &96.03 & 87.62& \textbf{89.90}&89.42\\
\hline
Blond\_Hair & 95.73 & \textbf{96.20} &96.08 &96.85 & 96.65&\textbf{97.04}\\
\hline
Brown\_Hair & 88.39 & 89.40 &\textbf{89.58}& 73.45& 76.59&\textbf{78.54}\\
\hline
Heavy\_Makeup & 91.21 & \textbf{91.90}&91.87 & 92.86& 93.87&\textbf{93.99}\\
\hline
No\_Beard & 96.19 & 96.46 &\textbf{96.52} & 80.00& 77.99&\textbf{80.52}\\
\hline
Oval\_Face & 74.64 & \textbf{76.57} &76.38 & 71.61& 71.96&\textbf{73.69}\\
\hline
Pointy\_Nose & 75.63 & 77.73 &\textbf{77.79} & 76.49& 79.85&\textbf{82.13}\\
\hline
Rosy\_Cheeks & 94.80 & \textbf{95.42} &95.37 & 81.55& 85.49&\textbf{85.51}\\
\hline
Wavy\_Hair & 84.55 & \textbf{85.91} &85.62 & 76.25& 78.48&\textbf{80.24}\\
\hline
Lipstick & 93.97 & 94.10 & \textbf{94.20} & 92.68& 92.71&\textbf{93.20}\\
\hline
Young  & 87.66 & \textbf{89.18} &88.70 & 85.42& 85.68&\textbf{85.98}\\
\hline
Gray\_Hair & 98.03 & 98.28 &\textbf{98.33} &88.51 & 88.44&\textbf{90.21}\\
\hline
Pale\_Skin & 96.68 & 96.97 &\textbf{97.17}& 83.87& 69.64&\textbf{90.86}\\
\hline
Blurry & 96.11 & 96.25 &\textbf{96.42 }& 83.87& 83.35&\textbf{85.18}\\
\hline
Black\_Hair & 89.14 & 89.92 &\textbf{90.29} & 90.00& 86.22&\textbf{90.49}\\
\hline
Straight\_Hair & 83.67 & 84.51 &\textbf{84.92} & 71.61& 68.54&\textbf{77.26}\\
\hline
Eyeglasses & 99.68 & 99.62 &\textbf{99.71} & 91.37& 88.33&\textbf{92.32}\\
\hline
Hat & 99.06 & 99.09 &\textbf{99.18} & 90.60& 85.98&\textbf{90.79}\\
\hline
5 O.C. Shadow  & 94.31 & 94.72 &\textbf{94.98} & 74.17& 70.10&\textbf{75.51}\\
\hline
Bald & 98.89 & 98.91 &\textbf{99.01} & 91.07& 89.14&\textbf{92.51}\\
\hline
Goatee & 97.13 & 97.41 &\textbf{97.63} & 79.94& 76.89&\textbf{81.40}\\
\hline
Male & 97.65 & 98.11 &\textbf{98.48} & \textbf{92.44}& 90.45&92.20\\
\hline
Mustache & 96.92 & 96.91 &\textbf{97.03} & 90.48& 87.05&\textbf{92.40}\\
\hline
Sideburns & 97.53 & 97.86 &\textbf{97.96} & 77.38& 73.36&\textbf{79.46}\\
\hline
Necktie & 97.05 & 97.06 & \textbf{97.08} & 80.30& 78.91&\textbf{80.79}\\
\hline
Arched\_Eyebrow & 81.66 & 83.15 &\textbf{83.89} & \textbf{78.21}& 75.68&78.18\\
\hline
Bags\_Under\_Eye & 84.60 & 84.85 &\textbf{85.46} & 78.45& 76.07&\textbf{79.30}\\
\hline
Big\_Lips & 70.06 & 71.36 &\textbf{71.76 }& 72.68& 68.20&\textbf{74.48}\\
\hline
Big\_Nose & 83.78 & 84.41 &\textbf{84.56} & 80.06& 77.05&\textbf{82.02}\\
\hline
Bushy\_Eyebrow & 92.20 & 92.29 &\textbf{92.80} & 70.30& 68.47&\textbf{75.24}\\
\hline
Chubby & 95.22 & \textbf{95.81} &95.76 & 73.93& 73.26&\textbf{74.73}\\
\hline
Double\_Chin & 95.89 & \textbf{96.42} &\textbf{96.42}& \textbf{79.46}& 75.71&79.17\\
\hline
Narrow\_Eyes & 85.75 & 86.94 &\textbf{87.48} & 77.26& 72.35&\textbf{78.05}\\
\hline
Recede\_Hair & 93.70 & 93.72 &\textbf{93.82} & 83.75& 82.78&\textbf{84.11}\\
\hline
Earring & 90.14 & 90.46 &\textbf{91.04} & 92.74& 91.96&\textbf{93.17}\\
\hline
Necklace & 86.69 & 87.71 &\textbf{88.15} & 88.15& 87.90&\textbf{88.81}\\
\hline
\hline
Average & 90.90 & 91.45 & \textbf{91.66}  &82.50 & 80.85&\textbf{84.34}\\
\hline
\end{tabular}
}
\end{table*}
\subsection{Results on the Multi-label Learning Network}
In this experiment, we evaluate the performance of MNet using 40 attributes, as done in \cite{liu2015deep12}. The detailed setting of MNet is given as follows: Firstly, MNet is initialized based on FNet. Then, we use 162,770 images on the CelebA dataset to train MNet. Finally, we test MNet with 19,962 images on the CelebA dataset. For the LFWA dataset, we use 6,571 images to train MNet and use 6,571 images to evaluate its performance.

\begin{figure*}[!t]
\centering
\subfloat[The CelebA dataset]{\includegraphics[width=4.8in]{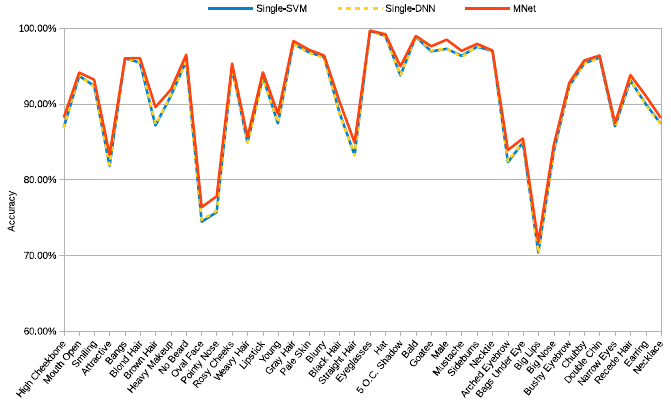}}
\hfil
\subfloat[The LFWA dataset]{\includegraphics[width=4.8in]{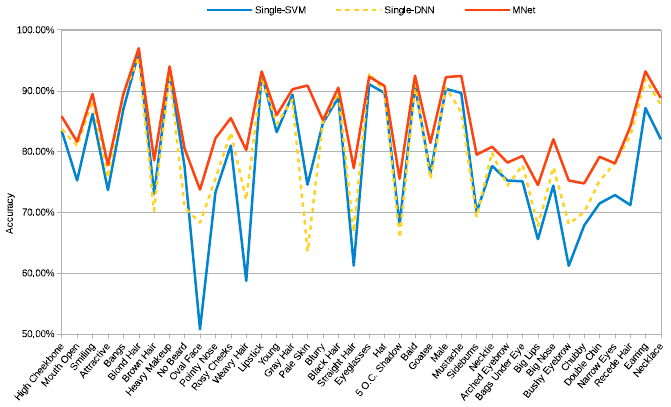}}
\caption{Performance comparison between the two methods based on  single-label learning and MNet based on  multi-label learning on  (a) the CelebA dataset and (b) the LFWA dataset.}
\label{fig:4}
\end{figure*}
\subsubsection{Influence of Loss Weight}
In this subsection, we evaluate the performance of MNet with different loss weight schemes on the CelebA and LFWA datasets, respectively. In Table 1, we show the classification accuracy obtained by MNet with different loss weight schemes, where MNet with the same loss weight scheme (i.e., the loss weight for each attribute is set to 1.0) and MNet with the proposed loss weight scheme are represented by `MNet\_Eq' and `MNet\_Prop', respectively. In `MNet\_Emp', we emphasize the attributes of one group among all groups (i.e., the attributes in one group are assigned with higher loss weights than those in the other groups). In particular, in this experiment we choose one group containing multiple attributes (e.g., the group 2 shown in Fig.~1) and assign the large loss weight (i.e., 1.0) to each attribute in this group. We assign the small loss weight (i.e., 0.1) to each attribute in the other different groups. \\
\indent From the experimental results in Table 1, we can see that  `MNet\_Prop' achieves better results than `MNet\_Eq' on both CelebA and LFWA datasets. In comparison to `MNet\_Eq', the improvement of  `MNet\_Prop' on the LFWA dataset is more obvious (about~2\% improvement). The improvement is more significant for some individual attributes (for example, `MNet\_Prop' achieves about~4\% higher classification accuracy for the `Attractive' attribute with regard to `MNet\_Eq' on the LFWA dataset ). The proposed `MNet\_Prop' effectively exploits the correlation between facial attributes by taking advantage of attribute grouping. More specifically, the loss weights corresponding to different facial attributes are assigned according to the results of attribute grouping (see Eq.~(4)). In this manner,  `MNet\_Prop' balances the loss weights for different facial attributes, which not only considers the differences among the facial attributes (note that `MNet\_Eq' treats each facial attribute equally), but also avoids over-fitting to one group (note that `MNet\_Emp' tends to focus on training one group, leading to the problem of over-fitting).\\
\indent Note that the performance of facial attribute classification is greatly affected by large facial appearance variations caused by pose, illumination, occlusion, etc. However, some facial attributes are more sensitive to the facial appearance variations than the other facial attributes. In other words, the classification task for some facial attributes is more difficult than that for the other facial attributes. For example, the variations caused by the face pose have more influence on the `Pointy\_Nose' attribute than the `Gray\_Hair' attribute. Thus, the classification performance obtained by the proposed method on the `Gray\_Hair' attribute (about 98\% classification accuracy on the CelebA dataset) is much better than that on the `Pointy\_Nose' attribute (about 78\% classification accuracy on the CelebA dataset). 

\subsubsection{Single-label Learning vs. Multi-label Learning}
Secondly, we compare the  proposed MNet with two methods (Single-SVM and Single-DNN) that are based on single-label learning (each attribute is learnt separately) on the CelebA and LFWA datasets, respectively. Specifically, Single-SVM denotes the method that uses DNN to learn the features of facial attributes, and then applies an SVM classifier for classification. `Single-DNN' represents the method that uses DNN to extract facial features and predict facial attributes. The classification accuracy comparison is reported in Fig.~4. \\
\indent From Fig.~4, we can observe that the proposed MNet achieves better results than the two competing methods that are based on single-label learning, which validates the effectiveness of the proposed method. 
This is because that multi-label learning effectively exploits the correlation between different facial attributes. In contrast, the network based on single-label learning does not consider the relationship among the facial attributes. Therefore, the single-label learning based methods do not make full use of the intrinsic information of facial attributes. \\
\indent In summary, compared with the single-label learning, the multiple attributes can be simultaneously exploited in MNet. Furthermore, the proposed loss weight scheme assigns different loss weights for different attributes, which makes MNet based on multi-label learning more effective than the network based on single-label learning.

\subsubsection{Comparison with the State-of-the-art Methods}
Finally, we compare the performance of MNet with the following state-of-the-art facial attribute classification methods trained with the labelled information: FaceTracer \cite{kumar2008facetracer33}, two versions of PANDA \cite{zhang2014panda11} (i.e., PANDA-w and PANDA-l), two versions of LNets+ANet \cite{liu2015deep12} (i.e., LNets+ANet (w/o) (without pre-training) and LNets+ANet), and MT-RBM (PCA) \cite{ehrlich2016facial}. FaceTracer \cite{kumar2008facetracer33} uses the features of HOG and the color histogram to train an SVM for facial attribute classification. PANDA \cite{zhang2014panda11} employs hundreds of poselets, which are aligned to predict facial attributes. LNets + ANet \cite{liu2015deep12} cascades two deep neural networks to detect the face and then learns the facial attributes from the detected parts. MT-RBM (PCA) \cite{ehrlich2016facial} learns the joint feature representation of the faces and facial landmark points for predicting facial attributes.\\
\begin{table*}[!t]
\caption{The classification accuracy (\%) comparison between MNet and several state-of-the-art methods for 40 facial attributes on the CelebA dataset and the LFWA dataset. The best results are in boldface.}
\center
\scalebox{0.8}{
\begin{tabular}{c|c|c|c|c|c|c|c|c|c|c|c|c|c|c|c}
\hline
\multirow{3}{*}{Attributes} &
\multicolumn{8}{c|}{CelebA} &
\multicolumn{7}{|c}{LFWA}\\
\cline{2-16}
 & \rotatebox{90}{FaceTracer} & \rotatebox{90}{PANDA-w} & \rotatebox{90}{PANDA-l}  & \rotatebox{90}{ANet} & \rotatebox{90}{LNets+ANet(w/0)}  & \rotatebox{90}{LNets+ANet}  &\rotatebox{90}{ MT-RBM (PCA)} &\rotatebox{90}{MNet} &\rotatebox{90}{FaceTracer} &  \rotatebox{90}{PANDA-w}  & \rotatebox{90}{PANDA-l}  & \rotatebox{90}{ANet} & \rotatebox{90}{LNets+ANet(w/0)} & \rotatebox{90}{LNets+ANet} & \rotatebox{90}{MNet} \\
  & \cite{kumar2008facetracer33} & \cite{zhang2014panda11} & \cite{zhang2014panda11} & \cite{li2013learning} & \cite{liu2015deep12} & \cite{liu2015deep12} & \cite{ehrlich2016facial} &  & \cite{kumar2008facetracer33} & \cite{zhang2014panda11} & \cite{zhang2014panda11} & \cite{li2013learning}& \cite{liu2015deep12} & \cite{liu2015deep12}  & \\
\hline
High\_Cheekbone & 84 & 80 & 86 & 85 & 84 & 87 & 83
& \textbf{88.19} & 77 & 75 & 86 & 79 & 83 & \textbf{88} &85.79\\
\hline
Mouth\_Open & 87 & 82 & 93 & 85 & 86 & 92 & 82
& \textbf{94.16} & 77 & 74 & 78 & 76 & 78 & \textbf{82} &81.59\\
\hline
Smiling & 89 & 89 & 92 & 92 & 88 & 92 & 88
& \textbf{93.21} & 78 & 77 & 89 & 82 & 88 & \textbf{91} &89.49\\
\hline
Attractive & 78 & 77 & 81 & 79 & 77 & 81 & 76
& \textbf{83.29} & 71 & 70 & 81 & 75 & 80 & \textbf{83} &77.78\\
\hline
Bangs & 88 & 89 & 92 & 94 & 92 & 95 & 88
& \textbf{96.03} & 72 & 79 & 84 & 84 & 84 & 88 &\textbf{89.42} \\
\hline
Blond\_Hair & 80 & 81 & 93 & 86 & 91 & 95 & 91
& \textbf{96.08} & 88 & 87 & 94 & 90 & 94 & 97 &\textbf{97.04}\\
\hline
Brown\_Hair & 60 & 69 & 77 & 74 & 78 & 80 & 83
& \textbf{89.58} & 62 & 65 & 74 & 71 & 73 & 77 &\textbf{78.54}\\
\hline
Heavy\_Makeup & 85 & 84 & 90 & 87 & 85 & 90 & 85
& \textbf{91.87} & 88 & 86 & 93 & 89 & 91 & \textbf{95} &93.99\\
\hline
No\_Beard & 90 & 87 & 93 & 91 & 92 & 95 & 90
& \textbf{96.52} & 69 & 63 & 75 & 69 & 75 & 79 &\textbf{80.52}\\
\hline
Oval\_Face & 64 & 62 & 65 & 65 & 63 & 66 & 73
& \textbf{76.38} & 66 & 64 & 72 & 66 & 71 & \textbf{74} &73.69\\
\hline
Pointy\_Nose & 68 & 65 & 71 & 67 & 70 & 72 & 73
& \textbf{77.79} & 74 & 68 & 76 & 72 & 76 & 80 &\textbf{82.13}\\
\hline
Rosy\_Cheeks & 84 & 81 & 87 & 85 & 87 & 90 & 94
& \textbf{95.37} & 70 & 64 & 73 & 71 & 72 & 78 &\textbf{85.51}\\
\hline
Wavy\_Hair & 73 & 76 & 77 & 79 & 75 & 80 & 72
& \textbf{85.62} & 62 & 63 & 75 & 65 & 73 & 76 &\textbf{80.24}\\
\hline
Lipstick & 89 & 88 & 93 & 91 & 90 & 93 & 89
& \textbf{94.20} & 87 & 83 & 93 & 86 & 92 & \textbf{95} &93.20\\
\hline
Young  & 80 & 77 & 84 & 81 & 83 & 87 & 81
& \textbf{88.70} & 80 & 76 & 82 & 79 & 82 & \textbf{86} &85.98\\
\hline
Gray\_Hair & 90 & 88 & 94 & 93 & 93 & 97 & 97
& \textbf{98.33} & 78 & 77 & 81 & 82 & 81 & 84 &\textbf{90.21}\\
\hline
Pale\_Skin & 83 & 84 & 91 & 89 & 87 & 91 & 96
& \textbf{97.17} & 70 & 64 & 84 & 68 & 81 & 84 &\textbf{90.86}\\
\hline
Blurry & 81 & 77 & 86 & 83 & 80 & 84 & 95
& \textbf{96.42} & 73 & 70 & 74 & 75 & 70 & 74 &\textbf{85.18}\\
\hline
Black\_Hair & 70 & 74 & 85 & 77 & 84 & 88 & 76
& \textbf{90.29} & 76 & 78 & 87 & 82 & 86 & 90 &\textbf{90.49}\\
\hline
Straight\_Hair & 63 & 67 & 69 & 70 & 69 & 73 & 80
& \textbf{84.92} & 67 & 68 & 73 & 72 & 71 & 76 &\textbf{77.26}\\
\hline
Eyeglasses & 98 & 94 & 98 & 96 & 96 & 99 & 96
& \textbf{99.71} & 90 & 84 & 89 & 88 & 92 & \textbf{95} &92.32\\
\hline
Hat & 89 & 91 & 96 & 93 & 96 & 99 & 97
& \textbf{99.18} & 75 & 78 & 82 & 82 & 84 & 88 &\textbf{90.79}\\
\hline
5 O.C. Shadow & 85 & 82 & 88 & 86 & 88 & 91 & 90
& \textbf{94.98} & 70 & 64 & 84 & 78 & 81 & \textbf{84} & 75.51\\
\hline
Bald & 89 & 92 & 96 & 92 & 95 & 98 & 98
& \textbf{99.01} & 77 & 82 & 84 & 86 & 83 & 88 &\textbf{92.51}\\
\hline
Goatee & 93 & 86 & 93 & 92 & 92 & 95 & 96
& \textbf{97.63} & 69 & 65 & 75 & 68 & 75 & 78 &\textbf{81.40}\\
\hline
Male & 91 & 93 & 97 & 95 & 94 & 98 & 90
& \textbf{98.48} & 84 & 86 & 92 & 91 & 91 & \textbf{94} &92.20\\
\hline
Mustache & 91 & 83 & 93 & 87 & 91 & 95 & 97
& \textbf{97.03} & 83 & 77 & 87 & 79 & 87 & 92 &\textbf{92.40}\\
\hline
Sideburns & 94 & 90 & 93 & 94 & 91 & 96 & 96
& \textbf{97.96} & 71 & 68 & 76 & 72 & 72 & 77 &\textbf{79.46}\\
\hline
Necktie & 86 & 88 & 91 & 90 & 86 & 93 & 94
& \textbf{97.08} & 71 & 70 & 79 & 72 & 76 & 79 &\textbf{80.79}\\
\hline
Arched\_Eyebrow & 76 & 73 & 78 & 75 & 74 & 79 & 77
& \textbf{83.89} & 67 & 63 & 79 & 66 & 78 & \textbf{82} &78.18\\
\hline
Bags\_Under\_Eye & 76 & 71 & 79 & 77 & 73 & 79 & 81
& \textbf{85.46} & 65 & 63 & 80 & 72 & 79 & \textbf{83} &79.30\\
\hline
Big\_Lips & 64 & 61 & 67 & 63 & 66 & 68 & 69
& \textbf{71.76} & 68 & 64 & 73 & 70 & 72 & \textbf{75} &74.48\\
\hline
Big\_Nose & 74 & 70 & 75 & 74 & 75 & 78 & 81
& \textbf{84.56} & 73 & 71 & 79 & 73 & 76 & 81 &\textbf{82.02}\\
\hline
Bushy\_Eyebrow & 80 & 76 & 86 & 80 & 85 & 90 & 88
& \textbf{92.80} & 67 & 63 & 79 & 69 & 79 & \textbf{82} &75.24\\
\hline
Chubby & 86 & 82 & 86 & 86 & 86 & 91 & 95
& \textbf{95.76} & 67 & 65 & 69 & 68 & 70 & 73 &\textbf{74.73}\\
\hline
Double\_Chin & 88 & 85 & 88 & 90 & 88 & 92 & 96
& \textbf{96.42} & 70 & 64 & 75 & 70 & 74 & 78 &\textbf{79.17}\\
\hline
Narrow\_Eyes & 82 & 79 & 84 & 83 & 77 & 81 & 86
& \textbf{87.48}  & 73 & 68 & 73 & 74 & 77 & \textbf{81} &78.05\\
\hline
Recede\_Hair & 76 & 82 & 85 & 84 & 85 & 89 & 92
& \textbf{93.82} & 63 & 61 & 84 & 70 & 81 & \textbf{85} &84.11\\
\hline
Earring & 73 & 72 & 78 & 77 & 78 & 82 & 81
& \textbf{91.04} & 88 & 85 & 92 & 87 & 90 & \textbf{94} &93.17\\
\hline
Necklace & 68 & 67 & 67 & 70 & 68 & 71 & 87
& \textbf{88.15} & 81 & 79 & 86 & 81 & 83 & 88 &\textbf{88.81}\\
\hline
\hline
Average  & 81 & 79 & 85 & 83 & 83 & 87 & 87
& \textbf{91.66} & 74 & 71 & 81 & 76 & 79 & \textbf{84} & \textbf{84.34}\\
\hline
\end{tabular}
}
\end{table*}
\indent Classification accuracies obtained by all the competing methods are reported in Table 2. For the CelebA dataset, we compare MNet with all the competing methods. The results obtained by the competing methods on the CelebA dataset are taken from \cite{ehrlich2016facial}. For the LFWA dataset, we compare MNet with all the competing methods except for MT-RBM (PCA) \cite{ehrlich2016facial}, because that MT-RBM (PCA) \cite{ehrlich2016facial} does not release the code on the LFWA dataset. Moreover, the LFWA dataset is not suitable for MT-RBM (PCA) to train and test, since the scale of the LFWA dataset is small and the data distribution of the LFWA dataset is unbalanced. The results obtained by the competing methods on the LFWA dataset are taken from \cite{liu2015deep12}. We follow the same evaluation protocol provided in \cite{liu2015deep12,ehrlich2016facial}.\\
\indent As shown in Table 2, MNet significantly outperforms the other competing methods on the CelebA dataset. Moreover, MNet achieves similar average accuracy compared with LNet + ANet (and obtains much better results than FaceTracer, PANDA-w and PANDA-l) on the LFWA dataset. LNet + ANet uses multiple patches cropped from the face region for data augmentation, which generates much more training data than our proposed method (using half of the images on the LFWA dataset). Thus, the training complexity of LNet + ANet \cite{liu2015deep12} is much higher than that of MNet. In general, MNet achieves better or comparable performance compared with the state-of-the-art methods.
\subsection{Results of the Proposed Method}
In this section, we evaluate the performance of the proposed FMTNet method for transfer learning on the CelebA dataset. The reason why we do not evaluate the proposed FMTNet on the LFWA dataset is that the data distribution of this dataset is unbalanced (for example, almost all the images are labelled with the `Male' attribute and the `Pointy\_Nose' attribute, while only a few images are labelled with the `Heavy\_Makeup' attribute), making the LFWA dataset difficult to be used for evaluating the performance of transfer learning.
\subsubsection{Direct Transfer}
Firstly, we perform the direct transfer method from one attribute to other attributes on the CelebA dataset. In other words, we train a model with the labelled data of one facial attribute in the source domain, and then we use the trained model to directly predict other facial attributes in the target domain. The classification accuracy obtained on one attribute, which is directly transferred to other attributes on the CelebA dataset is reported in Fig.~5. Specifically, we use 162,770 images to train the `Attractive' attribute which has the labelled information (while other attributes can also be used, and we observe similar results). Then, we predict the other attributes with this trained model. \\
\indent As shown in Fig.~5, because the model is trained with the labelled information of the `Attractive' attribute, the best results are obtained in predicting the `Attractive' attribute. Furthermore, the accuracy obtained on  the three attributes -- `Young', `Lipstick' and `Heavy\_Makeup' -- is slightly better than that obtained on the other attributes, because these three attributes belong to the same group as the `Attractive' attribute. However, the results for other transfer tasks are worse, since we train the model without using the labelled information of these attributes.
It is worth pointing out that although the `No\_Beard', `Oval\_Face' and `Pointy\_Nose' attributes belong to the same group as the `Attractive' attribute, the classification accuracy obtained on these attributes is not high. Therefore, the direct transfer method is not desirable and the performance is not stable.
\begin{figure}[!t]
\centering
\includegraphics[width=3.4in]{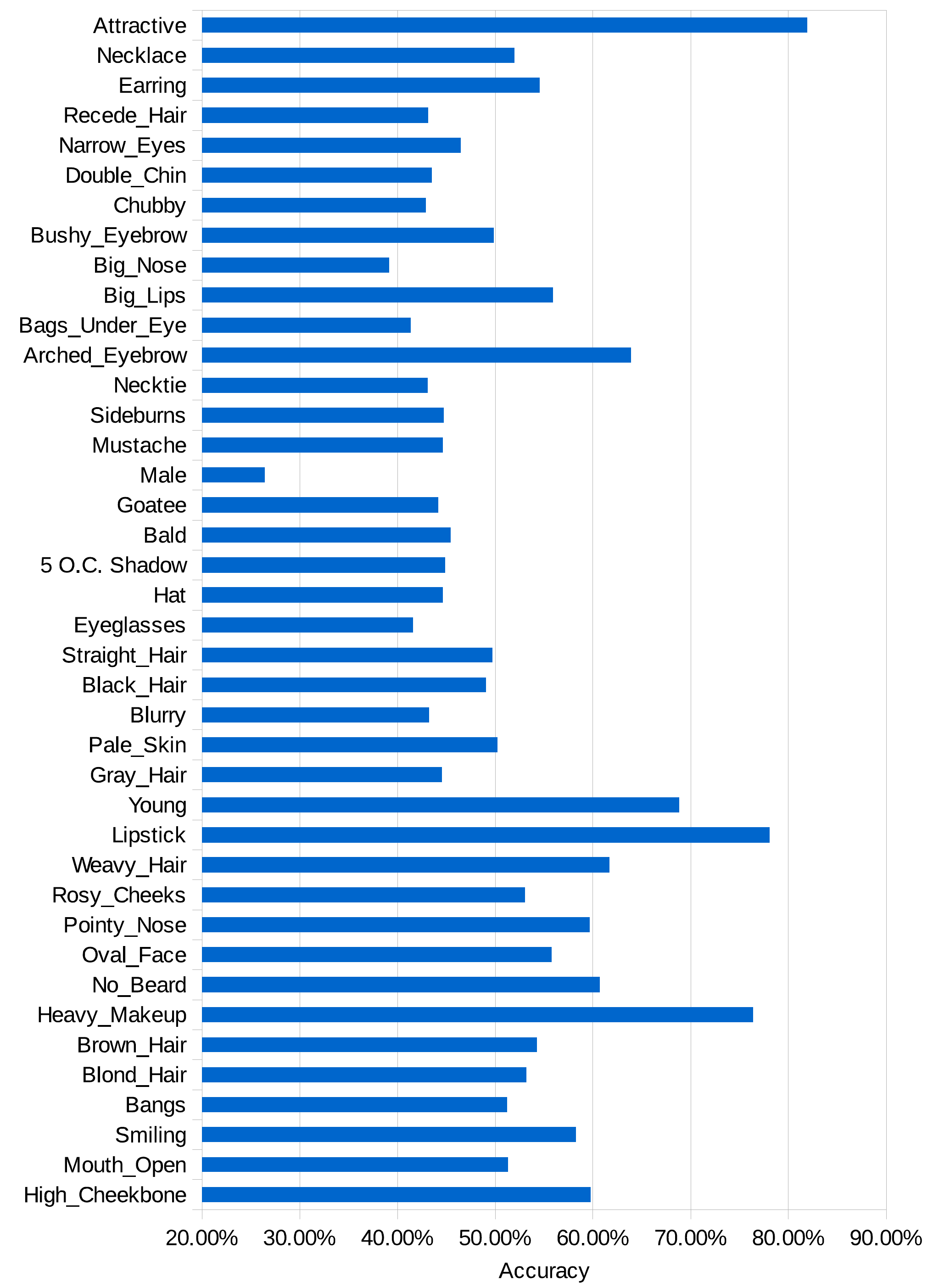}
\caption{Classification accuracy obtained on different attributes when we directly transfer one attribute to the other attributes. We train one attribute (`Attractive') and directly use the trained model to predict all the 40 attributes on the CelebA dataset.}
\label{fig:5}
\end{figure}

\begin{figure}[!t]
\centering
\includegraphics[width=3.4in]{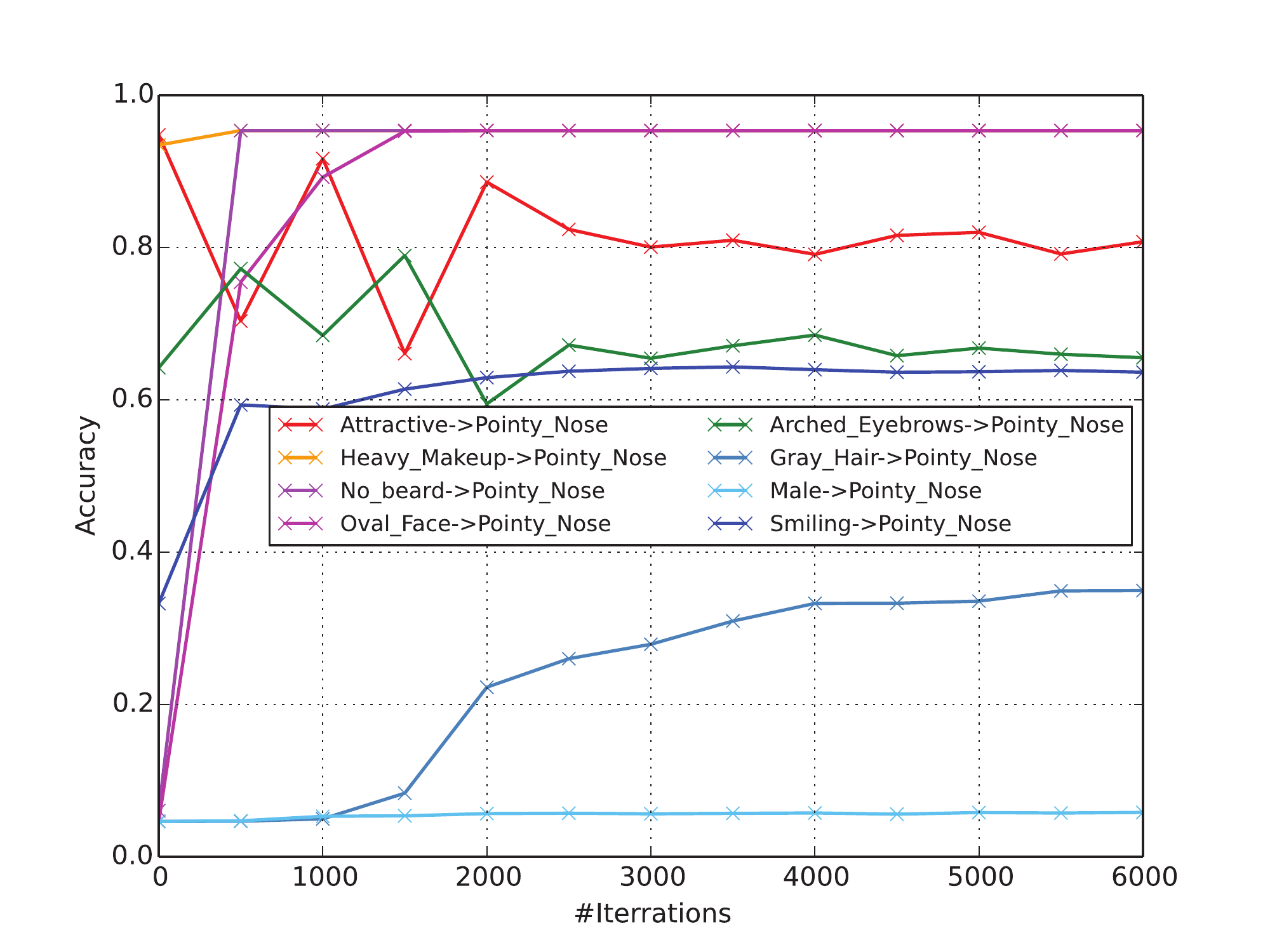}
\caption{Performance comparison on different source domains transferring to the same target domain during the training process. The source domains have different correlations with the target domain. Note that the loss weight $\alpha$ of the softmax loss in  TNet is set to 1.0 in this experiment.}
\label{fig:6}
\end{figure}
\subsubsection{Transfer under Different Correlations}
The correlation between the source and target domains will greatly affect the performance of unlabelled facial attribute classification. In this subsection, we evaluate the performance of FMTNet under different correlations between the source and target domains.\\
\indent We use 162,770 images as the labelled source data, and 19,962 images as the unlabelled target data on the CelebA dataset. Then, we use FMTNet, which transfers the source data to the target data for facial attribute classification. In this experiment, we randomly select eight labelled attributes (`Attractive', `Heavy\_Makeup',  `No\_Beard', `Oval\_Face', `Arched\_Eyebrow', `Gray\_Hair', `Male' and `Smiling') as the source data, and unlabelled attribute `Pointy\_Nose' as the target data (other attributes can also be used to obtain similar results). Therefore, we build the following eight transfer learning tasks: {`Attractive'} $\rightarrow$ {`Pointy\_Nose'}; {`Heavy\_Makeup'} $\rightarrow$ {`Pointy\_Nose'}; {`No\_Beard'} $\rightarrow$ {`Pointy\_Nose'}; {`Oval\_Face'} $\rightarrow$ {`Pointy\_Nose'}; {`Arched\_Eyebrow'} $\rightarrow$ {`Pointy\_Nose'}; {`Gray\_Hair'} $\rightarrow$ {`Pointy\_Nose'}; {`Male'} $\rightarrow$ {`Pointy\_Nose'}; {`Smiling'} $\rightarrow$ {`Pointy\_Nose'}. \\
\indent As mentioned in Section 3.3, the total loss of the overall TNet is the combination of the softmax loss and three MK-MMD losses. In this experiment, to explore the influence of the correlation between the source and target domains in FMTNet, the loss weight $\alpha$ for the softmax loss (see Eq.~(9)) is set to 1.0.\\
\indent The five attributes (`Attractive', `Heavy\_Makeup',  `No\_Beard', `Oval\_Face' and `Pointy\_Nose') are clustered into the same group (see Fig.~1). Therefore, these attributes have high correlation. On the other hand, the four attributes (`Arched\_Eyebrow', `Gray\_Hair', `Male' and `Smiling') and the `Pointy\_Nose' attribute belong to different groups \cite{liu2015deep12}. In other words, these attributes have low correlation to each other. The performance comparison on different transfer learning tasks over the training process is reported in Fig.~6.\\
\indent  From Fig.~6, the proposed FMTNet obtains better performance for the four transfer tasks ({`Attractive'} $\rightarrow$ {`Pointy\_Nose'}, {`Heavy\_Makeup'} $\rightarrow$ {`Pointy\_Nose'}, {`No\_Beard'} $\rightarrow$ {`Pointy\_Nose'} and
{`Oval\_Face'} $\rightarrow$ {`Pointy\_Nose'}) than for the other four transfer tasks ({`Arched\_Eyebrow'} $\rightarrow$ {`Pointy\_Nose'}, {`Gray\_Hair'} $\rightarrow$ {`Pointy\_Nose'}, {`Male'} $\rightarrow$ {`Pointy\_Nose'} and  {`Smiling'} $\rightarrow$ {`Pointy\_Nose'}). This is mainly because the features from the source attribute and the target attribute are similar when they belong to the same group. Therefore, FMTNet achieves better performance when the correlation between the source attribute and the target attribute is high. In contrast, when the source and target attributes belong to different groups, the performance of the transfer learning task drops because of the significant difference of the data distributions between the two attributes. In summary, the more correlated the two attributes are, the better the transferring performance is (when the value of $\alpha$ is fixed to 1.0).\\
\indent It is worth pointing that the red curve (i.e., `Attractive' $\rightarrow$ `Pointy\_Nose') is not rather stable at the initial 3000 iterations compared with the other curves. This is mainly because that the imbalanced training data are used at the beginning of the whole training process. However, the training accuracy becomes stable as the number of iterations increases, since the classification capability of CNN is gradually enhanced with the number of iterations. Recall that the number of training data for each transfer task is the same. But some facial attributes have many labelled data, while some other facial attributes only have a small number of labelled data. For example, there are totally 4,000 training data for each facial attribute, where 2,500 images are labelled as `Heavy\_Makeup', while only 1,000 images are labelled as `Attractive'. The small number of labelled data in the source domain (for example, in the task of `Attractive' $\rightarrow$ `Pointy\_Nose') will cause that the training accuracy is not stable at the beginning of the training process. In contrast, the training accuracy for some transfer tasks (for example, in the task of `Heavy\_Makeup' $\rightarrow$ `Pointy\_Nose') is more stable since the labelled data in the source domain are sufficient. Actually, Rudd \textit{et al.}~\cite{rudd2016moon} have mentioned the problem of imbalanced data for facial attribute classification. 
\subsubsection{Comparison with the State-of-the-art Methods}
In this subsection, we compare the performance of the proposed FMTNet with several state-of-the-art transfer learning methods, including TCA \cite{pan2011domain34}, MIDA \cite{yan2016domain35}, ITL \cite{shi2012information36} and GFK \cite{gong2012geodesic6}. TCA \cite{pan2011domain34} uses the MMD-regularized PCA, which is a conventional transfer learning method, to learn some transfer components across domains. MIDA \cite{yan2016domain35} reduces the discrepancy between the source and target domains by maximizing the independence between the source and target features. ITL \cite{shi2012information36} consistently learns a domain-invariant feature space and optimizes an information-theoretic metric on the target domain. GFK \cite{gong2012geodesic6} bridges the source and target domains by interpolating the intermediate subspace. \\
\indent We use 5,000 images with the labelled information on the CelebA dataset as the source data, and 2,000 images without the labelled information on the CelebA dataset as the target data (also as the test data). The facial attributes are randomly chosen as the source data and the target data. For the traditional transfer learning methods, we extract the features from the source data and target data with single-label learning and FNet, respectively. Then, we use the logistical regression technique to predict the facial attribute.\\
\indent Firstly, we choose the eight attributes (i.e., `Attractive', `Heavy\_Makeup',  `No\_Be\-ard', `Oval\_Face', `Arched\_Eyebrow', `Gray\_Hair', `Male' and `Smiling') as the source data and the `Pointy\_Nose' attribute as the target data. According to the experimental results in Section 4.3.2, the performance of FMTNet is significantly affected by the correlation between the source and target attributes. Specifically, when the source and target attributes belong to the same group, FMTNet can achieve excellent results (with $\alpha=1$). FMTNet uses the MK-MMD loss to measure the difference between the source domain and the target domain, which makes it more discriminative than the other transfer learning methods. However, when the source and target attributes are from different groups, the mean accuracy obtained by FMTNet becomes worse (with $\alpha=1$). This is mainly due to the fact that when the attributes belong to different groups, the large value of $\alpha$ makes the trained model easily overfit to the source attribute. Therefore, when the source and target attributes belong to different groups, we also set the value of $\alpha$ to be 0.1. In other words, the network focuses more on the minimization of the three MK-MMD losses (mainly reducing the domain discrepancy), when the value of $\alpha$ is small. Therefore, FMTNet uses the MK-MMD loss to measure the difference between the source domain and the target domain, which makes it more discriminative than the other transfer learning methods. The mean accuracy obtained by all the competing methods is shown in Fig.~7, where the loss weight $\alpha$  of the softmax loss is set to 1.0 or 0.1. The `Same Group' represents the attributes (i.e., `Attractive', `Heavy\_Makeup',  `No\_Beard' and `Oval\_Face') belonging to the same group as the `Pointy\_Nose' attribute. The `Different Groups' represents the attributes (i.e., `Arched\_Eyebrow', `Gray\_Hair', `Male' and `Smiling') belonging to different groups compared with the `Pointy\_Nose' attribute.\\
\indent As shown in Fig.~7, the mean accuracy obtained by FMTNet outperforms that obtained by the competing state-of-the-art methods. Specifically, if the source and target attributes belong to the same group, the mean accuracy obtained by FMTNet achieves the top performance when the value of $\alpha$ is set to 1.0, as shown in Fig.~7 (a). And if the source and target attributes belong to  different groups, the mean accuracy obtained by FMTNet achieves the top performance when the value of $\alpha$ is set to 0.1, as shown in Fig.~7 (b). In summary, the loss weight $\alpha$ should be chosen according to the relationship between two attributes for better performance. More specifically, if the source and target attributes belong to the same group, the loss weight $\alpha$ should be assigned with a large value. Otherwise, $\alpha$ should be assigned with a small value. \\
\indent To show the feature transferability of the proposed method, Fig.~8 demonstrates the t-SNE embeddings \cite{Donahue2014} of the samples for the {`Attractive'} $\rightarrow$ {`Pointy\_\-Nose'} task with the original features and the transferred features (on the source and target domains), respectively. We observe that, with the original features, the data distributions on the source and target domains are quite different. However, with the transferred features obtained by the FMTNet method, the data distributions on the two domains become more similar. This observation further verifies the effectiveness of MK-MMD, which significantly reduces the domain discrepancy.\\
\indent To further demonstrate the effectiveness of the proposed FMTNet method, we randomly choose 20 facial attributes as the source data and 20 facial attributes as the target data under two different cases (i.e., the attributes in the source domain and in the target domain are from the same group or from different groups). The performance comparison among all the competing methods is given in Table 3. As shown in Table 3, the proposed FMTNet (we set $\alpha=1$ when the source and  target attributes belong to the same group and $\alpha=0.1$ when the source and target attributes belong to different groups) obtains the top mean accuracy among all the competing methods. Therefore, the loss weight $\alpha$ should be chosen according to the results of attribute grouping (i.e., whether the source and target attributes belong to the same group or not).
\begin{figure*}[!t]
\centering
\subfloat[Same Group]{\includegraphics[width=2.3in]{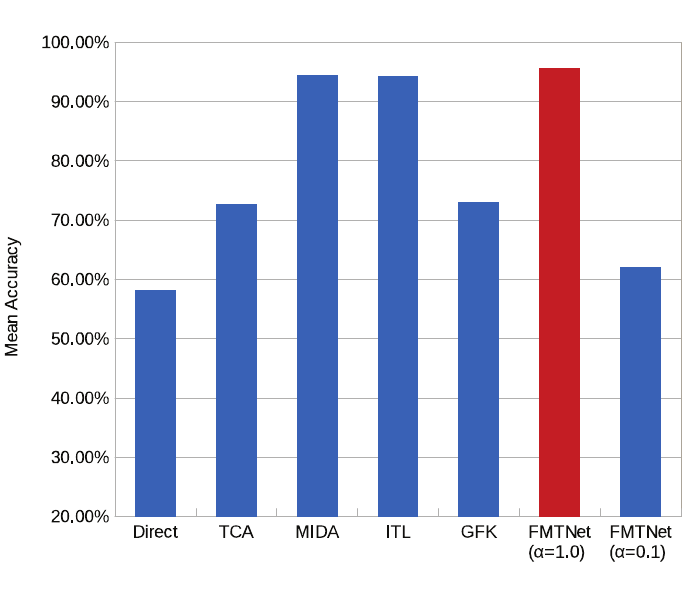}}
\hfil
\subfloat[Different Groups]{\includegraphics[width=2.3in]{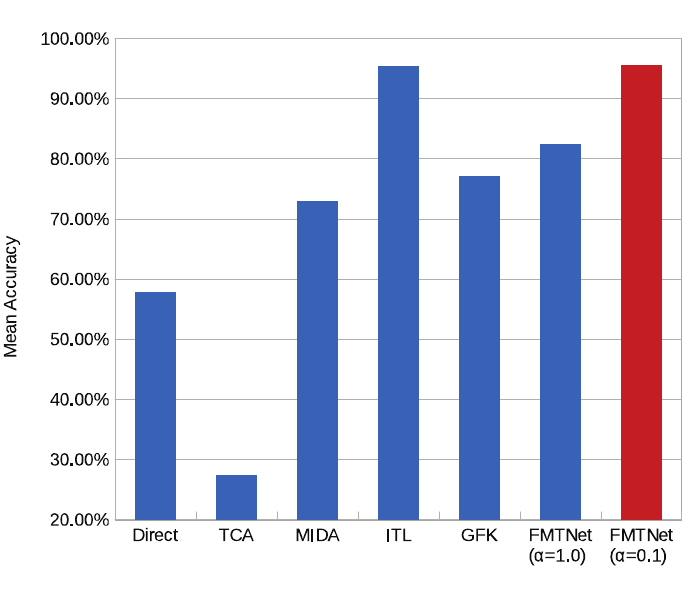}}
\caption{The mean accuracy (\%) obtained by the proposed FMTNet and five other state-of-the-art methods on the eight attributes (i.e., `Attractive', `Heavy\_Makeup',  `No\_Beard', `Oval\_Face', `Arched\_Eyebrow', `Gray\_Hair', `Male' and `Smiling') transferring to the `Pointy\_Nose' attribute obtained by the proposed FMTNet and six other state-of-the-art methods. (a) represents the results obtained for the four attributes (i.e., `Attractive', `Heavy\_Makeup',  `No\_Beard' and `Oval\_Face') belonging to the same group as the `Pointy\_Nose' attribute. (b) represents the results obtained for the other four attributes (i.e., `Arched\_Eyebrow', `Gray\_Hair', `Male' and `Smiling') belonging to different groups from the one of the `Pointy\_Nose' attribute.}
\label{fig:7}
\end{figure*}
\begin{figure*}[!t]
\centering
\subfloat[The original features]{\includegraphics[width=2.4in,height=2.8in]{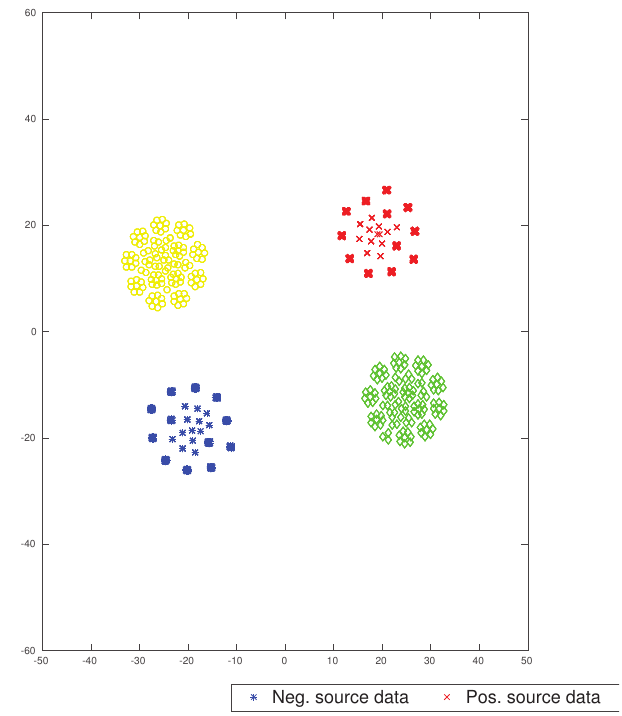}}
\subfloat[The transferred features]{\includegraphics[width=2.2in,height=2.8in]{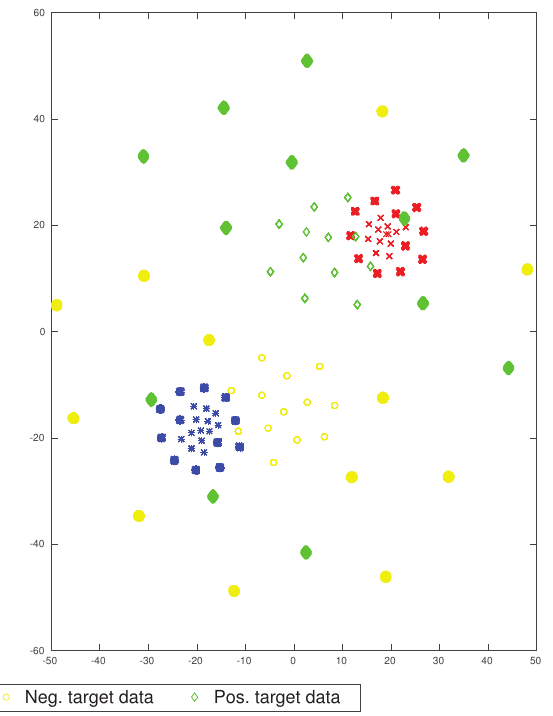}}
\caption{Feature visualization: t-SNE of (a) the original features and (b) the transferred features obtained by FMTNet for {`Attractive'} $\rightarrow$ {`Pointy\_Nose'} on the source and target domains.}
\label{fig:8}
\end{figure*}
\begin{table*}[!t]
\caption{The mean accuracy (\%) obtained by the proposed FMTNet in comparison with the state-of-the-art methods. The best results are in boldface.}
\center
\scalebox{0.8}{
\centerline{\begin{tabular}{c|c|c|c|c|c|c|c}
\hline
 \multirow{2}{*}{Group} & \multirow{2}{*}{Methods} & \multirow{2}{*}{Direct} & TCA & MIDA & ITL & GFK & \multirow{2}{*}{FMTNet}  \\
 & & & \cite{pan2011domain34} & \cite{yan2016domain35} &  \cite{shi2012information36} & \cite{gong2012geodesic6} & \\
\hline
\multirow{20}{*}{Same}
  & High\_Cheekbone$\rightarrow$Mouth\_Open & 49.65& 49.75 & 49.85 & 49.95& 49.95 &\textbf{71.95}\\
\cline{2-8}
 \multirow{21}{*}{($\alpha$=1.0)}
  & Smiling$\rightarrow$High\_Cheekbone &35.60&49.20 & 49.05&49.35 &49.65& \textbf{74.75}\\
\cline{2-8}
  & Attractive$\rightarrow$Bangs &56.65& 16.00 & 15.85& 16.85 &47.75 & \textbf{82.90}\\
\cline{2-8}
  & Blond\_Hair$\rightarrow$Brown\_Hair &65.00&19.65 & 30.80 & 67.25 &73.50 & \textbf{81.30}\\
\cline{2-8}
  & Heavy\_Makeup$\rightarrow$No\_Beard &72.36&82.60 & \textbf{85.10} & 81.95 &28.45 & 75.49\\
\cline{2-8}
  & Pointy\_Nose$\rightarrow$Oval\_Face &65.20&29.45 & 57.75 &66.65 &64.25 & \textbf{71.15}\\
\cline{2-8}
  & Rosy\_Cheeks$\rightarrow$Wavy\_Hair &37.75&38.50 & 61.95 & \textbf{62.10} &59.80 & \textbf{62.10}\\
\cline{2-8}
  & Lipstick$\rightarrow$Young &75.10&75.50 & \textbf{76.00} & 75.30 &40.00 & 75.88\\
\cline{2-8}
  & Gray\_Hair$\rightarrow$Pale\_Skin &95.90&23.45 & 95.95 & 4.45 &92.75 & \textbf{99.05}\\
\cline{2-8}
  & Blurry$\rightarrow$Gray\_Hair &81.35&4.55 & 96.55 & 96.60 &93.30 &\textbf{99.65}\\
\cline{2-8}
  & Black\_Hair$\rightarrow$Straight\_Hair &78.75&21.00 & 21.05 & 70.20 &68.00 & \textbf{79.30}\\
\cline{2-8}
  & Eyeglasses$\rightarrow$ Hat &5.08&12.20 & \textbf{95.70} & 95.60 &93.80 & 95.61\\
\cline{2-8}
  & 5 O.C. Shadow$\rightarrow$Bald &57.51&3.45 & 94.95 & \textbf{97.60} &90.45 & \textbf{97.60}\\
\cline{2-8}
  & Goatee$\rightarrow$Male &36.05&39.50 & 61.55& 60.45 &\textbf{63.50} & 61.45\\
\cline{2-8}
  & Mustache$\rightarrow$Sideburns &95.41&5.90 & \textbf{95.45} & 94.15 &93.25 & \textbf{95.45}\\
\cline{2-8}
  & Arched\_Eyebrow$\rightarrow$Bags\_Under\_Eye &30.00&20.45 & 40.40 & 65.10 &73.45 & \textbf{80.40}\\
\cline{2-8}
  & Big\_Lips$\rightarrow$Big\_Nose &77.35&21.55 & 73.00 &78.50 &77.15 & \textbf{78.90}\\
\cline{2-8}
  & Bushy\_Eyebrow$\rightarrow$Chubby &88.18&7.20 & 79.65 & 80.70 &88.20 & \textbf{94.15}\\
\cline{2-8}
   & Double\_Chin$\rightarrow$Narrow\_Eyes &21.05&27.60 &85.10 & 85.10 &82.60 &\textbf{85.20}\\
\cline{2-8}
   & Recede\_Hair$\rightarrow$Earring &21.15&21.80 & 78.60 & \textbf{79.00} &75.15 & \textbf{79.00}\\
\cline{2-8}
\specialrule{0em}{1pt}{1pt}
\cline{2-8}
  & Average &57.25&28.47 & 67.22 &68.84 &70.25 & \textbf{82.06}\\
\hline
\hline
\multirow{20}{*}{Different}
 & High\_Cheekbone$\rightarrow$Attractive &47.00& 49.90 & 49.90 & 49.60 &41.30&\textbf{51.10} \\
\cline{2-8}
\multirow{21}{*}{($\alpha$=0.1)}
 & Mouth\_Open$\rightarrow$Bangs &61.70& 15.75 &15.90 &16.55 &44.70 & \textbf{84.40}\\
\cline{2-8}
 & Smiling$\rightarrow$Blond\_Hair &57.85& 13.10 &12.55 &13.25 &44.35 & \textbf{87.05}\\
 \cline{2-8}
 & Gray\_Hair$\rightarrow$Brown\_Hair &77.15& 32.90 &81.20 &19.00 &78.30 & \textbf{81.25}\\
 \cline{2-8}
 & Pale\_Skin$\rightarrow$Heavy\_Makeup &41.85& 41.60 &58.70 &\textbf{58.80} &57.15 & 58.64\\
 \cline{2-8}
 & Blurry$\rightarrow$No\_Beard &85.90& 84.70 &14.00 &14.15 &15.45 & \textbf{86.04}\\
 \cline{2-8}
 & Black\_Hair$\rightarrow$Oval\_Face &67.70& 29.25 &29.30 &64.75 &60.75 & \textbf{71.15}\\
 \cline{2-8}
 & Straight\_Hair$\rightarrow$Pointy\_Nose &36.40& 30.60 &56.40 &60.20 &61.95 & \textbf{69.90}\\
 \cline{2-8}
 & Eyeglasses$\rightarrow$Rosy\_Cheeks &8.85& 15.40 &92.60 &\textbf{92.70} &89.70 & 90.63\\
 \cline{2-8}
 & Hat$\rightarrow$Wavy\_Hair &42.95& 38.55 &62.05 &38.55 &61.05 & \textbf{62.10}\\
 \cline{2-8}
 & 5 O.C. Shadow$\rightarrow$Arched\_Eyebrow &30.70& 30.95 &67.15 &\textbf{69.70} &63.65 & 68.80\\
 \cline{2-8}
 & Bald$\rightarrow$Bags\_Under\_Eye &78.75& 26.05 &\textbf{80.40} &20.50 &79.95 & \textbf{80.40}\\
 \cline{2-8}
 & Goatee$\rightarrow$Big\_Lips &54.80& 35.85 &\textbf{65.20} &64.00 &64.35 & 64.10\\
  \cline{2-8}
 & Male$\rightarrow$Big\_Nose &73.90 & 23.50 &21.70 &22.50 &64.05 & \textbf{78.86}\\
  \cline{2-8}
 & Mustache$\rightarrow$Bushy\_Eyebrow &60.40& 14.00 &87.25 &86.15 &85.15 & \textbf{87.26}\\
 \cline{2-8}
 & Sideburns$\rightarrow$Chubby &72.25& 8.75 &94.10 &6.60 &92.05 &\textbf{94.15} \\
 \cline{2-8}
 & Necktie$\rightarrow$Double\_Chin &5.70& 6.25 &\textbf{94.70} &5.90 &91.55 & 94.68\\
 \cline{2-8}
 & Lipstick$\rightarrow$Narrow\_Eyes &48.25& 15.55 &15.05 &15.65 &56.15 & \textbf{63.50}\\
 \cline{2-8}
 & Young$\rightarrow$Recede\_Hair &43.70& 9.75 &10.20 &9.80 &24.50 &\textbf{63.85} \\
 \cline{2-8}
 & Earring$\rightarrow$Bangs &82.50 & 15.80 &33.55 &64.20 &72.75 & \textbf{83.90}\\
\cline{2-8}
\specialrule{0em}{1pt}{1pt}
\cline{2-8}
  & Average &53.92&26.91 & 52.10 & 39.63 &62.44 & \textbf{76.09}\\
\hline
\end{tabular}}
}
\end{table*}

The proposed FMTNet method takes advantage of some existing components. However, our method is not simply stacked by these components. In FMTNet, three carefully designed networks (i.e., FNet, MNet and TNet) are used, where these networks share the same structure at their former layers and they differ at their latter layers. Therefore, the networks can be effectively trained via fine-tuning. In other words, TNet can be effectively fine-tuned based on MNet (which is initialized by FNet). Especially, a loss weight scheme is proposed to explicitly exploit the correlation between facial attributes based on attribute grouping, which can improve the generalization performance of the proposed method.\\
\indent In general, compared with the other transfer learning methods, the proposed FMTNet, which transfers features from the source domain to the target domain, is very effective. The reasons are summarized as follows: \\
\indent (1) Instead of directly performing transfer learning for facial attribute classification, we take advantage of multi-label learning to learn the network parameters (MNet) and then fine-tune the transfer network (TNet) based on MNet. In other words, we use a hierarchical training strategy, where TNet is fine-tuned based on MNet, and MNet is fine-tuned based on FNet. Therefore, TNet and MNet can effectively trained via fine-tune. \\
\indent (2)  We improve existing multi-label learning and transfer learning components by exploiting the correlation between facial attributes and propose a loss weight scheme based on attribute grouping.

\section{Conclusions}
For the task of facial attribute classification, we have presented a novel method, termed FMTNet, which consists of three different sub-networks -- FNet, MNet and TNet -- for face detection, multi-label learning and transfer learning, respectively. MNet in FMTNet predicts multiple facial attributes simultaneously for the labelled facial attributes. Moreover, MNet reduces feature redundancy with the proposed loss weight scheme, resulting in significant performance improvements. Based on MNet, TNet in FMTNet predicts facial attributes with unlabelled information by using MK-MMD. The proposed method, which combines multi-label learning with transfer learning, is general and can be applied to other computer vision tasks.

\section*{Acknowledgments}
This work was supported by the National Natural Science Foundation of China under Grants 61571379, 61503315, U1605252, and 61472334, by the Natural Science Foundation of Fujian Province of China under Grant 2017J01127, by the Fundamental Research Funds for the Central Universities under Grant 20720170045, and by the National Key Research and Development Program of China under Grant 2017YFB1302400.

\end{document}